\documentclass[runningheads]{llncs}
\usepackage{graphicx}
\usepackage{background}

\SetBgContents{The final authenticated version is available online at https://doi.org/[insert DOI].}
\SetBgScale{1}
\SetBgAngle{0}
\SetBgOpacity{1}
\SetBgColor{black}
\SetBgPosition{current page.south west}
\SetBgHshift{8cm}
\SetBgVshift{1cm}
\begin{document}
\title{From web crawled text to project descriptions: automatic summarizing of social innovation projects}
\titlerunning{Automatic summarization of social innovation}
%

\author{Nikola Milo\v{s}evi\'{c}\\
  School of Computer Science,\\ University of Manchester\\
  {\tt \small nikola.milosevic@manchester.ac.uk}  \And
  Dimitar Marinov\\
  School of Computer Science,\\ University of Manchester\\
  {\tt \small dimitar.marinov@student.manchester.ac.uk}  \And
 Goran Nenadi\'{c}\\
  School of Computer Science,\\ University of Manchester\\
  {}}
\author{Nikola Milo\v{s}evi\'{c}\inst{1}\orcidID{0000-0003-2706-9676} \and
Dimitar Marinov\inst{1}\orcidID{0000-0001-6197-9679} \and
Abdullah G\"{o}k\inst{2}\orcidID{0000-0002-9378-3336} \and
Goran Nenadi\'{c}\inst{1}\orcidID{0000-0003-0795-5363}}
\authorrunning{N. Milosevic et al.}
%
\institute{School of Computer Science, University of Manchester, M13 9PL, Manchester, UK
\email{nikola.milosevic@manchester.ac.uk}
\email{dimitar.marinov@student.manchester.ac.uk}
\and 
Hunter Centre For Entrepreneurship, Strathclyde Business School, University of Stratclyde, Glasgow, UK
}
\maketitle              
\begin{abstract}
In the past decade, social innovation projects have gained the attention of policy makers, as they address important social issues in an innovative manner. A database of social innovation is an important source of information that can expand collaboration between social innovators, drive policy and serve as an important resource for research. Such a database needs to have projects described and summarized. In this paper, we propose and compare several methods (e.g. SVM-based, recurrent neural network based, ensambled) for describing projects based on the text that is available on project websites. 
We also address and propose a new metric for automated evaluation of summaries based on topic modelling.

\keywords{Summarization  \and evaluation metrics \and text mining \and natural language processing \and social innovation \and SVM \and neural networks}
\end{abstract}

\section{Introduction}

Social innovations are projects or initiatives that address social issues and needs in an innovative manner \cite{bonifacio2014social}. In the past decade, social innovation has gained significant attention from policy makers and funding agencies around the worlds, especially in the EU, USA, and Canada. Policy makers and researchers are particularly interested in monitoring social innovation projects, the effects of policies on these projects and the effects of these projects for the society. 

In order to enable monitoring of social innovation projects a number of database creation projects were funded over time. In the KNOWMAK project, we aim to integrate and expand on previously collected information by utilizing automation approaches enabled by machine learning and natural language processing techniques. 

The existing data sources for social innovation are varied in their levels of depth and detail. Therefore, in KNOWMAK we aim to normalize the information, providing the same wealth of information for each reported project. In order to do this, we utilize the data from original data sources, as well as the data from the projects' webpages and social media sites, such as Facebook and Twitter. 

In order to provide relevant information to the researchers and policy makers, the projects in the database need to be described. Some of the original data sources have descriptions, but many data sources do not have. Additionally, some of the descriptions in existing data sources may be too long (e.g. over 500 words), or too short (1 sentence) and therefore need to be normalized. 

Automated summarization can be used to automate and speed up the process of summarizing texts about a project in the database. Summarization is a well-known task in natural language processing, however solutions in literature do not address the domain specific issues. Project description building using summarization has challenges that may not be present with a usual text summarization task. In this task, it is necessary to generate short, cohesive description that best portrays the project, which may be described over several web pages, contain noisy text (pages or portions of pages with irrelevant text) and align project description to the theme of the database. 

In this paper, we compare several methods for creating project descriptions and summaries in the semi-automated system that takes texts about social innovation projects from the web. We develop a method that makes human readable project descriptions from the scraped pages from the project web sources. This paper presents an automated project description method applied in the KNOWMAK project that aims to create a tool for mapping knowledge creation in the European area. The project focuses on collecting information on publications, patents, EU projects and social innovation projects. As publications, patents and EU projects would have abstracts or short descriptions, this paper aims at the particular case of describing social innovation projects.

\section{Background}

Automatic summarization is a complex natural language processing task which has been approached from several perspectives. We will review the main approaches. 

On the whole, it is challenging to evaluate automatic summarization. Summaries of text will look different depending on who is doing them and which approach is used. However, it has to be ensured that the main points of the text that is analysed have been retained. Over the years, there have been a couple of evaluation metrics proposed. In this section, we will also review the proposed metrics. 

\subsection{Summarization approaches}
Summarization approaches can be classified into two main categories: (1) extractive and (2) abstractive \cite{nallapati2017summarunner}. Extractive approaches try to find snippets, sentences and paragraphs that are important, while abstractive approaches attempt to paraphrase important information from the original text. The types of summarizers may also depend on how many documents are used as input (single-document or multi-document), on the languages of input and output (monolingual, multilingual or cross-lingual), or purpose factors (informative, indicative, user-oriented, generic or domain specific) \cite{dong2018survey}. 

Summarization approaches can be both supervised and unsupervised. Unsupervised methods usually use sentence or phrase scoring algorithms to extract the relevant parts of the original text \cite{riedhammer2010long,fattah2009ga}. Most of the extractive summarization approaches model the problem as a classification task, classifying whether certain sentences should be included in the summary or not \cite{sinha2018extractive}. These approaches usually use graphs, linguistic scoring or machine learning in order to classify sentences. Standard machine learning classifiers, such as Naive Bayes or Support Vector Machines (SVM) using features such as the frequency of words \cite{sarkar2011using,bazrfkan2014using,neto2002automatic}, as well as neural network-based classifiers \cite{nallapati2017summarunner,sinha2018extractive,dong2018survey} have been proposed. Traditional machine learning classifiers usually use features such as the frequency of phrases, relational ranks, positions of the sentences in the text, or overlapping rate with the text title. Neural network approaches utilize word, sentence and document representations as vectors, pre-trained on large corpora (word, document or sentence embeddings). Then these vectors are imputed into convolutional or recurrent neural networks for classification training. 

Abstractive summarization is considered less traditional \cite{young2018recent}. Approaches usually include neural network architectures trained on both original texts and human created summaries. Approaches using sequence-to-sequence neural architectures \cite{nallapati2016abstractive}, but also attention mechanism have been proposed \cite{rush2015neural}.

\subsection{Evaluation measures for summarization}

A good summary should be a short version of the original text, carrying the majority of relevant content and topics in condensed format. Summarization of a text is a subjective problem for humans and it is hard to define what a good summary would consist of. However, a number of quantitative metrics have been proposed, such as ROUGE or Pyramid. 

Recall-Oriented Understudy for Gisting Evaluation (ROUGE) is a commonly used metric in summarization literature \cite{dong2018survey} that is based on overlapping n-grams in summary and original text. There are several variants of ROUGE, such as ROUGE-N (computing percentage of the overlapping n-grams), ROUGE-L (computing the longest overlapping N-gram), ROUGE-S (computing the overlapping skip-grams in the sentence) \cite{lin2004rouge}. Since ROUGE takes into account only overlapping n-grams, it often favors the summaries that are long, where the summarizer did not sufficiently reduced the size of the original text. 

Pyramid is another metric that is based on the assumption that there is no one best summary of the given original text \cite{nenkova2004evaluating}. Pyramid requires a number of human generated summaries for each text as well as human annotations for summarization content units (SCU). For each SCU a weight is assigned based on the number of human generated summaries containing it. Newly created summaries are evaluated based on the overlapping SCUs and their weights. This method is expensive, since it requires a lot of human labour for annotating and generating multiple summaries for evaluated texts \cite{dong2018survey}. 

While ROUGE and Pyramid metrics are the most used in current literature, other approaches have been proposed. A Latent Semantic Analysis-based metric was proposed based on the hypothesis that the analysis of semantic elements of the original text and summary will provide a better metric about the portion of important information that is represented in the summary \cite{steinberger2012evaluation}. As ROUGE metrics often do not correlate with human rankings, the evidence was provided that LSA based metric correlates better than ROUGE and cosine similarity metric based on the most significant terms or topics. 

Human ranking and scoring is a measure that is often used for evaluation of summarization systems \cite{steinberger2012evaluation}. Human annotations are more expensive than automatic annotations, however, they provide a good metric that accounts for all elements of a good summary definition (main topics, condensed length, readability).

\section{Method}
\subsection{Method overview}

We present a comparison and implementation of four summarization or description generation methods for social innovation. The input to all summarization methods is text crawled from the social innovation project websites, while the expected output is a short and condensed description of the project (summary). 

\begin{figure}[h!]
	\centering
	\includegraphics[width=5in]{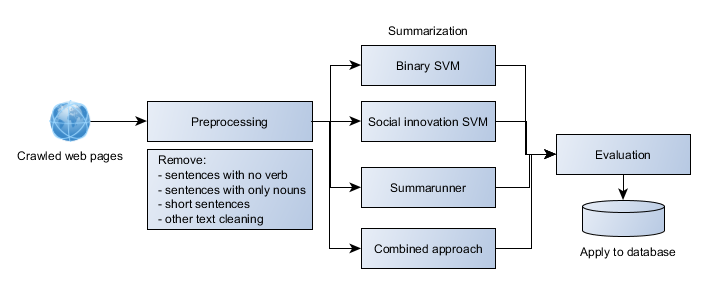}
	\caption{Methodology overview}
	\label{fig:method}
\end{figure}

The method consists of data collection, training data set generation, data cleaning, classification and evaluation steps. Figure \ref{fig:method} presents the methodology overview.

\subsection{Data collection and data set generation}

The initial set of social innovation projects was collected using existing databases of social innovation, such as MOPACT, Digital Social Innovation, InnovAge, SI-Drive, etc. The data was collected from a compiled list of about 40 data sources. Some of the data sources contained data that can be downloaded in CSV, JSON or XML format, however many data sources contained data accessible only through the website and therefore needed to be crawled. As these data sources contained structured data, with humanly created descriptions of the projects, websites and social media, a set of crawlers were created that were able to locate these structured data points on the page and store them in our database. Only a small number of data sources already contained descriptions of the projects and they were used for the creation of the training set. 

We collected 3560 projects. Out of these, 2893 project had identifiable websites. In order to provide data for describing the projects, we created a crawler that collects text from the websites. 

We performed a set of annotation tasks in which annotators were annotating sentences that describe how each project satisfies some of the following social innovation criteria: 
\begin{itemize}
    \item \textit{Social objective} - project addresses certain (often unmet) societal needs, including the needs of particular social groups; or aims at social value creation.
    \item \textit{Social actors and actor interactions} - involves actors who would not normally engage in innovation as an economic activity, including formal (e.g. NGOs, public sector organisations etc.) and informal organisations (e.g. grassroots movements, citizen groups, etc.) or creates collaborations between "social actors", small and large businesses and the public sector in different combinations
    \item \textit{Social outputs} - creates socially oriented outputs/outcomes. Often these outputs go beyond those created by conventional innovative activity (e.g. products, services, new technologies, patents, and publications), but conventional outputs/outcomes might also be present.
    \item \textit{Innovativeness} - There should be a form of "implementation of a new or significantly improved product (good or service), or process, a new marketing method, or a new organisational method".
\end{itemize}
Data annotation is further explained in \cite{milosevic2018classification}. The data set contained 315 documents, 43 of which were annotated by 4 different annotators, while the rest were mainly single annotated. The distribution of annotated sentences is presented in Table \ref{table:dataset}. Annotated data, descriptions from the original data sources and crawled websites were used for training and evaluating summarization approaches. 

\begin{table}[h!]
\centering
\begin{tabular}{ | l r | }
  \hline
  \small \textbf{Criteria} & \small \textbf{Number of sentences} \\ \hline
  \multicolumn{2}{|l|}{Social innovation criteria} \\ 
  \small Objectives  & \small 374 \\ 
  \small Actors & \small 217\\  
  \small Outputs & \small 309  \\ 
  \small Innovativeness & \small 256  \\
  \small Not satisfying any criteria & \small 3167  \\ \hline 
  \multicolumn{2}{|l|}{Binary (inside/outside summary)} \\ 
  \small Inside  & \small 2459 \\ 
  \small Outside & \small 12962\\  \hline
\end{tabular} 
\caption{Number of sentences satisfying social innovation criteria}
\label{table:dataset}
\end{table}

\subsection{Data cleaning}
The data from the websites may be quite noisy, as the crawler was collecting all textual information, including menus, footers of the pages and at times advertisements. Additionally, many pages contained events and blog posts that were not relevant for describing the core of the project. Therefore, we have performed some data cleaning before proceeding with training the summarizers. 

In order to reduce the amount of irrelevant text in form of menus and footers, we have performed part of speech tagging and excluded sentences that did not contain verbs. 

For further summarization, only main pages, about pages and project description pages were used. In case the page was not in English it was translated using Google Translate. 

\subsection{SVM based summarizer}

The first summarization approach is based on the assumption that the task can be modelled as a classification task, where sentences would be classified as part of a summary or not. It was hypothesized that words in a sentence would indicate whether it describes the project (e.g. "project aims to...", "the goal of the project is to...", etc.). 

In order to create a training data set, we utilized projects that had both project description in the original data sources and crawled websites. Since the descriptions were created by humans, they usually cannot be matched with the sentences from the website. In order to overcome this issue, we generated sent2vec embedding vectors of the sentences in both the description and the crawled text \cite{pagliardini2018unsupervised}. We then computed cosine similarities between the sentences from the description and the ones from the crawled text. If the cosine similarity is higher than 0.8, the sentence is labeled as part of the summary, otherwise it is labeled as a sentence that should not be part of the summary. 

These sentences were used as training data for the SVM classifier. Before training we balanced the number of positive (sentences that should be part of the summary) and negative (sentences that should remain outside the summary) instances. The bag-of-words transformed to TF-IDF scores, the position of a sentence in the document (normalized to the score between 0-1) and keywords were used as features for the SVM classifier. The keywords are extracted using KNOWMAK ontology \cite{maynard2017ontologies,zhang2018adapted} API that for the given text returns grand societal challenge topics and a set of keywords that were matched for the given topic and text\footnote{\url{https://gate.ac.uk/projects/knowmak/}}.

\subsection{Social innovation criteria classifier}
The social innovation criteria classifier utilized an annotated data set. In this data set, sentences that were marked as explaining why a project satisfies any of the social innovation criteria (objectives, actors, outputs, innovativeness), were used as positive training instances for the SVM classifier. The classifier used a bag-of-words transformed to TF-IDF scores. 

\subsection{Summarunner}
Summarunner is an extractive summarization method developed by IBM Watson \cite{nallapati2017summarunner} that utilizes recurrent neural networks (GRU). If compared using ROUGE metrics, the algorithm outperforms state-of-the-art methods. The method visits sentences sequentially and classifies each sentence by whether or not it should be part of the summary. The method is using a 100-dimensional word2vec language model \cite{mikolov2013distributed}. The model was originally trained on a CNN/DailyMail data set \cite{cheng2016neural}. The social innovation data set that we have created was quite small and not sufficient for training a neural network model (about 350 texts compared to over 200,000 in DailyMail data). However, we performed a model fitting on our social innovation data set.

\subsection{Stacked SVM-based summarizer and Summarunner}
Our final summarization method was developed as a combination of SVM-based method and Summarunner. We have noticed that binary SVM model produces quite long summaries and may be efficient for initial cleaning of the text. Once the unimportant parts have been cleaned up by the SVM-based classifier, Summarunner shortens the text and generates the final summary. 

\section{Evaluation methodology}
In order to evaluate our methodologies and select the best performing model we used ROGUE metrics, human scoring and two topic-based evaluation methods. 

ROUGE metrics are the most popular and widely used summarization scoring approaches which were presented back in 2004 \cite{lin2004rouge,nallapati2017summarunner,dong2018survey}. As such, we are utilizing them as well. 

Since a good summary should include the most important topics from the original text, topic-related metrics can be devised. We have used two topic based metrics: one was based on KNOWMAK ontology and the proportion of matched topics related to EU defined Grand Societal Challenges\footnote{\url{https://ec.europa.eu/programmes/horizon2020/en/h2020-section/societal-challenges}} and Key Enabling Technologies\footnote{\url{http://ec.europa.eu/growth/industry/policy/key-enabling-technologies\_en}} in the original and summarized text. The other method was based on latent Dirichlet allocation (LDA) \cite{blei2003latent}. We have extracted 30 topics using LDA from merged corpus of original texts and summaries and then we have calculated the proportion of topics that match. In order to prevent favouring long summaries, we have normalized the scores, assuming that the perfect summary should be no longer than 25\% of the length of the original text (longer texts were penalized). 

\section{Evaluation and results}
The evaluation of summarization techniques is a challenging process, therefore, we have employed several techniques. 

Since SVMs classifiers are utilizing classification, we have calculated their precision, recall and F1-scores. These are measures commonly used for evaluating classification tasks. These metrics are calculated on a test (unseen) data set, containing 40 documents (286 sentences labeled as inside summary, 2014 sentences as outside). The results can be seen in Table \ref{table:classification}.

\begin{table}[h!]
\centering
\begin{tabular}{ | l r r r | }
  \hline
  \small \textbf{Classifier} & \small \textbf{Precision}  & \small \textbf{Recall} & \small \textbf{F1-score}\\ \hline
  \small Binary SVM  & \small 0.8601 & \small 0.7130 & \small 0.7594 \\ 
  \hline 
  \small Objectives SVM & \small 0.8423 & \small 0.5601 & \small 0.6226\\  
  \small Actors SVM & \small 0.8821 & \small 0.4687 & \small 0.5659  \\ 
  \small Innovativeness SVM & \small 0.8263 & \small 0.4456 & \small 0.5166  \\
  \small Outputs SVM & \small 0.8636 & \small 0.6284 & \small 0.7089  \\
  \hline
\end{tabular} 
\caption{Evaluation based on classification metrics (precision, recall and F1-score) for classification-based summarizers (binary and social innovation criteria-based)}
\label{table:classification}
\end{table}

The data set for training these classifiers is quite small, containing between 200-400 sentences. It is interesting to note that the criteria classifiers containing larger number of training sentences (compare Table \ref{table:dataset} and Table \ref{table:classification}), perform with a better F1-score (Objectives and Outputs). This indicates that scores can be improved by creating a larger data set. The classifiers perform with quite good precision, which means there are few false positive sentences (the majority of the sentences that end up in summary are correct).

Since ROUGE metrics are commonly used in summarization literature, we have evaluated all our summarization approaches with ROUGE 1, ROUGE 2 and ROUGE-L metrics. The evaluation was performed again on an unseen test set, containing 40 documents and their summaries. The results can be seen in Table \ref{table:rouge}.

\begin{table}[h!]
\centering
\begin{tabular}{ | l r r r | }
  \hline
  \small \textbf{Classifier} & \small \textbf{ROUGE 1}  & \small \textbf{ROUGE 2} & \small \textbf{ROUGE-L}\\ \hline
  \small Binary SVM  & \small 0.6096 & \small 0.5544 & \small 0.5553 \\ 
  \small Social innovation SVM & \small 0.6388 & \small 0.6140 & \small 0.5846\\  
  \small Summarunner & \small 0.6426 & \small 0.5788 & \small 0.5762  \\ 
  \small Binary SVM + Summarunner & \small 0.5947 & \small 0.5197 & \small 0.5279 \\
  \small Binary SVM + Summarunner Relative Length & \small 0.5496 & \small 0.4731 & \small 0.4668  \\
  \hline
\end{tabular} 
\caption{ROUGE scores for the developed summarization methodologies}
\label{table:rouge}
\end{table}

Summarunner has the best performance based on unigram ROUGE (ROUGE-1) score. However, the social innovation SVM-based summarizer performs better in terms of bigram ROUGE (ROUGE-2) and ROUGE-L score (measuring longest common token sequence). Based on these results, it is possible to conclude that a specifically crafted classifier for the problem will outperform a generic summarizer, even if it was trained only on a small data set. Stacked binary SVM and Summarunner perform worse than single summarizers on their own in terms of ROUGE. 

In order to further evaluate the methodologies used, we have used an LDA-based metric. The assumption behind using this approach was that a good summarizer would have a high number of topics in the summary/description and the original text matching. The results of the LDA topic similarity evaluation can be seen in Table \ref{table:topics}. 

\begin{table}[h!]
\centering
\begin{tabular}{ | l r | }
  \hline
  \small \textbf{Classifier} & \small \textbf{LDA Topic Similarity}\\ \hline
  \small Binary SVM  & \small 0.2703\\ 
  \small Social innovation SVM & \small 0.2485\\  
  \small Summarunner & \small 0.2398\\ 
  \small Binary SVM + Summarunner & \small 0.2683\\
  \hline
\end{tabular} 
\caption{LDA topic similarity scores for the developed summarization methodologies}
\label{table:topics}
\end{table}

The most matching topics are found with the binary SVM classifier. However, this classifier is also producing the longest summaries. Stacked SVM and Summarunner are performing similar matches with much shorter summaries being generated. 

The second topic-based approach utilizes topics about grand societal challenges and key-enabling technologies retrieved from the KNOWMAK topic-modelling tool. The results can be seen in Table \ref{table:gate_topics}.

\begin{table}[h!]
\centering
\begin{tabular}{ | l r | }
  \hline
  \small \textbf{Classifier} & \small \textbf{KNOWMAK Topic Similarity}\\ \hline
  \small Binary SVM  & \small 0.3725\\ 
  \small Social innovation SVM & \small 0.3625\\  
  \small Summarunner & \small 0.3025\\ 
  \small Binary SVM + Summarunner & \small 0.3025\\
  \hline
\end{tabular} 
\caption{Topic similarity evaluation using KNOWMAK ontology topics}
\label{table:gate_topics}
\end{table}

The binary SVM summarizer, followed by the social innovation summarizer are the best methodologies according to this metric. 

Finally, summaries were scored by human annotators. Human scorers were presented with an interface containing the original text and a summary for each of the three methods (binary SVM, social innovation SVM and Summarunner). For each of the summaries they could give a score between 0-5. In Table \ref{table:human_scores} are presented averaged scores made by the human scorers. We have also averaged the scores in order to account for document length. In order to do that we used the following formula:
$$LengthAveragedScore = \frac{docLen-summaryLen}{docLen}*human\_score$$

\begin{table}[h!]
\centering
\begin{tabular}{ | l | r | r | r | }
  \hline
  \small \textbf{Classifier} & \small \textbf{Number of ratings} & \small \textbf{Human Score}  & \small \textbf{Length averaged human score}\\ \hline
  \small Binary SVM & 23 & \small 2.7391 & \small 0.8647\\ 
  \small Social innovation SVM & 20  & \small 2.4500 & \small 1.6862\\  
  \small Summarunner & 22 & \small 2.0000 & \small 1.5110\\ 
  \hline
\end{tabular} 
\caption{Human scores for the developed summarization methodologies}
\label{table:human_scores}
\end{table}

The best human scores were for binary SVM. However, this classifier  excluded only a few sentences from the original text, and it was generally creating longer summaries. If the scores are normalized for length, the best performing summarizer was the one based on social innovation criteria, followed by Summarunner. At the time of the human scoring, the stacked approach consisting of binary SVM and Summarunner was not yet developed, so results for this approach are not available. 

We have used stacked (SVM+Summarunner) and social innovation classifier in order to generate summaries for our database. Stacked model was used as fallback, in case summary based on social innovation model was empty or contained only one sentence. The approach was summarizing and generating project descriptions where either the description was too long (longer than 1000 words), or was missing. The summarizer generated new summaries for 2186 projects.

\section{Conclusion}

Making project descriptions and summaries based on the textual data available on the internet is a challenging task. The text from the websites may be noisy, different length, and important parts may be presented in different pages of the website. In this paper, we have presented and compared several approaches for a particular problem of summarizing social innovation projects based on the information that is available about them on the web. The presented approaches are part of a wider information system, including the ESID database\footnote{\url{https://esid.manchester.ac.uk/}} and the KNOWMAK\footnote{\url{https://www.knowmak.eu/}} tool. Since these approaches make extractive summaries, they may not have connected sentences in the best manner, and therefore additional manual checks and corrections would be performed before final publication of the data. However, these approaches significantly speed up the process of generating project descriptions. 

Evaluating automatically-generated summaries remains a challenge. A good summary should carry the most important content, but also significantly shorten the text. Finding a balance between the content and meaning that was carried from original text to the summary and final length can be quite challenging. Most of the currently used measures in the literature do not account for the summary length, which may lead to biases towards longer summaries. There are a number of measure that we have used and proposed in this work. Often, it is not easy to indicate strengths and weaknesses of summarization approaches using single measures and using multiple measures may be beneficial.

Most of the current research presents summarization approaches for general use. Even though, these approaches can be used in specific domains and for specific cases (such as social innovation), our evaluation shows that approaches developed for a particular purpose perform better overall.  

Our evaluation indicated that it may be useful to combine multiple summarization approaches. Certain approaches can be used to clear the text, while the others may be used to further shorten the text by carrying the most important elements of the text. In the end, we used a combined approach for the production of the summaries in our system.

\section*{Acknowledgments}
The work presented in this paper is part of the KNOWMAK project that has received funding from the European Union's Horizon 2020 research and innovation programme under grant agreement No. 726992.

\bibliographystyle{splncs04}
\bibliography{ranlp2015}

\end{document}